\documentclass[letterpaper]{article} 
\usepackage{aaai2026}  
\usepackage{times}  
\usepackage{helvet}  
\usepackage{courier}  
\usepackage[hyphens]{url}  
\usepackage{graphicx} 
\usepackage{natbib}  
\usepackage{caption} 
\usepackage{algorithm}
\usepackage{algorithmic}
\usepackage{bbold}
\usepackage{amsmath}
\usepackage{amssymb} 
\usepackage{multirow}
\usepackage{booktabs}
\usepackage{newfloat}
\usepackage{listings}
\DeclareCaptionStyle{ruled}{labelfont=normalfont,labelsep=colon,strut=off} 
\floatstyle{ruled}
\newfloat{listing}{tb}{lst}{}
\floatname{listing}{Listing}
\title{Bridging Vision and Language for Robust Context-Aware Surgical Point Tracking: The VL-SurgPT Dataset and Benchmark}
\author{
    Rulin Zhou\textsuperscript{\rm 1,\rm 3,\rm 5}\equalcontrib,
    Wenlong He\textsuperscript{\rm 4}\equalcontrib,
    An Wang\textsuperscript{\rm 1}\equalcontrib,
    Jianhang Zhang\textsuperscript{\rm 4},
    Xuanhui Zeng\textsuperscript{\rm 4},\\
    Xi Zhang\textsuperscript{\rm 4},
    Chaowei Zhu,\textsuperscript{\rm 2},
    Haijun Hu,\textsuperscript{\rm 2}\thanks{Corresponding authors.},
    Hongliang Ren\textsuperscript{\rm 1,\rm 5}\footnotemark[2]
}
\affiliations{
    \textsuperscript{\rm 1}The Chinese University of Hong Kong\\
    \textsuperscript{\rm 2}Division of Gastrointestinal Surgery, Shenzhen People’s Hospital\\
    \textsuperscript{\rm 3}The University of Hong Kong\\
    \textsuperscript{\rm 4}Shenzhen University\\
    \textsuperscript{\rm 5}The Chinese University of Hong Kong, Shenzhen Research Institute\\
    
    zhourulin@connect.hku.hk,
    wa09@link.cuhk.edu.hk,
    hu.haijun@szhospital.com,
    hlren@ee.cuhk.edu.hk

}

\begin{document}

\maketitle

\begin{abstract}
Accurate point tracking in surgical environments remains challenging due to complex visual conditions, including smoke occlusion, specular reflections, and tissue deformation. While existing surgical tracking datasets provide coordinate information, they lack the semantic context necessary to understand tracking failure mechanisms.
We introduce \textbf{VL-SurgPT}, the first large-scale multimodal dataset that bridges visual tracking with textual descriptions of point status in surgical scenes.
The dataset comprises 908 \textit{in vivo} video clips, including 754 for tissue tracking (17,171 annotated points across five challenging scenarios) and 154 for instrument tracking (covering seven instrument types with detailed keypoint annotations). We establish comprehensive benchmarks using eight state-of-the-art tracking methods and propose \textbf{TG-SurgPT}, a text-guided tracking approach that leverages semantic descriptions to improve robustness in visually challenging conditions. 
Experimental results demonstrate that incorporating point status information significantly improves tracking accuracy and reliability, particularly in adverse visual scenarios where conventional vision-only methods struggle. By bridging visual and linguistic modalities, VL-SurgPT enables the development of context-aware tracking systems crucial for advancing computer-assisted surgery applications that can maintain performance even under challenging intraoperative conditions.
\end{abstract}

\begin{links}
    \link{Project}{https://szupc.github.io/VL-SurgPT/}
\end{links}

\begin{figure}[t]
\centering
  \includegraphics[width=\linewidth]{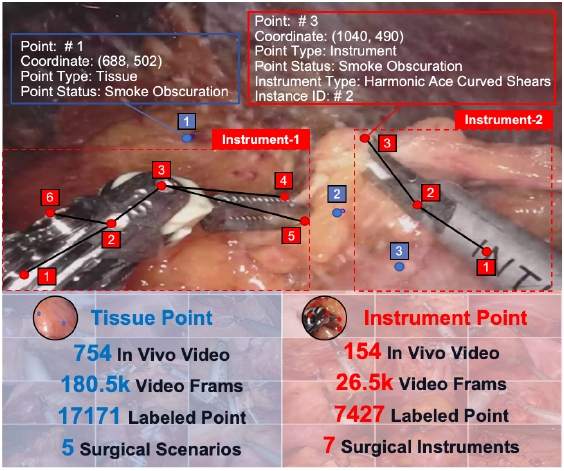}
  \caption{Overview of our Vision-Language Surgical Point Tracking (VL-SurgPT) dataset, a large-scale multimodal dataset containing visual and textual annotations for tissue and instrument points across diverse challenging scenarios.}
  \label{fig:teaser}
\end{figure}

\section{Introduction}

Accurate point tracking in surgical environments is critical for advancing computer-assisted interventions, enabling applications like motion understanding and scene perception~\cite{schmidt2024tracking,schmidt2024surgical,wu2025surgpose}. Unlike general computer vision scenarios where point tracking has achieved remarkable success~\cite{zheng2023pointodyssey,doersch2022tapvid}, surgical environments introduce extraordinary visual challenges, including dense electrocautery smoke, specular reflections from wet anatomical surfaces, dynamic instrument occlusions, and substantial tissue deformation. These challenges severely compromise the tracking performance of general tracking approaches~\cite{doersch2023tapir,doersch2024bootstap,karaev2024cotracker}.

Existing tracking datasets treat point tracking as purely geometric problems with visual annotations alone. General benchmarks like TAP-Vid~\cite{doersch2022tapvid} and PointOdyssey~\cite{zheng2023pointodyssey}, along with surgical-specific datasets like SurgT~\cite{cartucho2024surgt}, STIR~\cite{schmidt2024surgical}, and SurgPose~\cite{wu2025surgpose}, provide visual data but lack semantic context explaining why tracking fails under specific conditions. 

Furthermore, these datasets typically focus on either tissue or instrument tracking in isolation, failing to capture the holistic nature of surgical procedures where both elements interact dynamically. Recent efforts like SurgMotion~\cite{zhan2024surgmotion} began addressing this gap but remained unimodal, providing only coordinate annotations without semantic descriptions of visual conditions. 
This semantic gap represents a critical barrier to developing robust surgical tracking systems. Existing datasets overlook the point status where rich contextual information describing both the point's condition and the surrounding environmental state. We hypothesize that incorporating semantic descriptions of visual conditions can substantially enhance tracking performance by providing contextual guidance that pure visual methods cannot capture. 

To address this, we introduce \textbf{VL-SurgPT}, the first comprehensive multimodal dataset that bridges visual point tracking with semantic descriptions of environmental conditions. As shown in Fig.~\ref{fig:teaser}, our dataset combines coordinate annotations with synchronized textual descriptions characterizing each point's visual status, enabling fine-grained analysis of tracking behavior and developing context-aware methods.
Our contributions include: \textbf{(1)} \textbf{VL-SurgPT}, a large-scale multimodal dataset containing visual and textual annotations for surgical point tracking across diverse challenging scenarios; \textbf{(2)} comprehensive benchmarks evaluating eight tracking methods with condition-specific performance analysis; and \textbf{(3)} \textbf{TG-SurgPT}, a text-guided tracking approach that leverages semantic descriptions to achieve superior performance, particularly under adverse visual conditions where conventional methods struggle.

\section{Related Works}\label{sec:related}

\subsection{Surgical Point Tracking Datasets}

Existing surgical tracking datasets have primarily focused on either tissue points or instrument keypoints.
Early \textbf{tissue tracking} datasets, such as SuPer~\cite{li2020super} and Semantic SuPer~\cite{lin2023semantic}, were largely confined to \textit{ex vivo} settings and limited annotation quantities. SurgT~\cite{cartucho2024surgt} curated \textit{in vivo} surgical tissue tracking dataset but only focused on bounding box tracking rather than precise, continuous point trajectories. STIR~\cite{schmidt2024surgical} advanced the field by utilizing infrared fluorescent markers for more accurate point localization in both \textit{in vivo} and \textit{ex vivo} scenarios; however, its annotations were typically sparse and endpoint-focused. 
Regarding \textbf{instrument keypoint tracking} datasets, RMIT~\cite{sznitman2012data} and the EndoVis15 dataset~\cite{bodenstedt2018comparative} provided instrument keypoint labels, predominantly from \textit{ex vivo} setups with limited semantic depth regarding the visual status of the keypoints. SurgPose~\cite{wu2025surgpose} improved annotation efficiency for \textit{ex vivo} data using UV markers, primarily targeting 6DoF pose estimation rather than direct, continuous keypoint tracking. 
Recently, SurgMotion~\cite{zhan2024surgmotion} offered comprehensive annotations for tissue and instrument point tracking, yet it adhered to a unimodal, vision-only annotation paradigm.

A critical limitation across these datasets is their unimodal nature: while they provide geometric coordinates, they lack synchronized semantic descriptions of visual conditions (e.g., smoke, reflection, occlusion) that directly impact tracking performance in complex \textit{in vivo} surgical environments. This absence of multimodal information hinders the development and evaluation of tracking algorithms that can understand and adapt to these specific visual challenges. VL-SurgPT addresses this gap by offering rich textual annotations of point status in conjunction with precise spatial coordinates. 

\subsection{Surgical Point Tracking Methods}
Recent advances in point tracking like TAPIR~\cite{doersch2023tapir}, BootsTAP~\cite{doersch2024bootstap}, CoTracker~\cite{karaev2024cotracker, karaev2024cotracker3}, SEARAFT~\cite{wang2024sea}, MFT~\cite{neoral2024mft}, MFTIQ~\cite{serych2025mftiq}, and Track-On~\cite{Aydemir2025trackon} demonstrate strong performance on general tracking benchmarks like TAP-Vid~\cite{doersch2022tapvid}. 
Surgical-specific tracking methods have been developed to address domain challenges. Semantic SuPer~\cite{lin2023semantic} incorporated geometric and semantic cues for tissue tracking. SENDD~\cite{schmidt2023sendd} used graph neural networks for sparse keypoint matching. More recently, Ada-Tracker~\cite{guo2024ada} and Endo-TTAP~\cite{zhou2025endo} showed promising results on surgical datasets through adaptive matching and attention mechanisms, respectively. Besides, SurgMotion~\cite{zhan2024surgmotion} proposes to track surgical points of tissue and instruments simultaneously with mask and As Rigid As Possible (ARAP) constraints, and demonstrated strong performance on tissue tracking tasks. However, these methods still rely solely on visual cues and do not leverage semantic information to enhance robustness under challenging conditions.
This limitation motivates our approach of providing explicit visual status annotations to enable condition-specific analysis and method development.

\begin{figure*}[t]
  \centering
  \includegraphics[width=.92\linewidth]{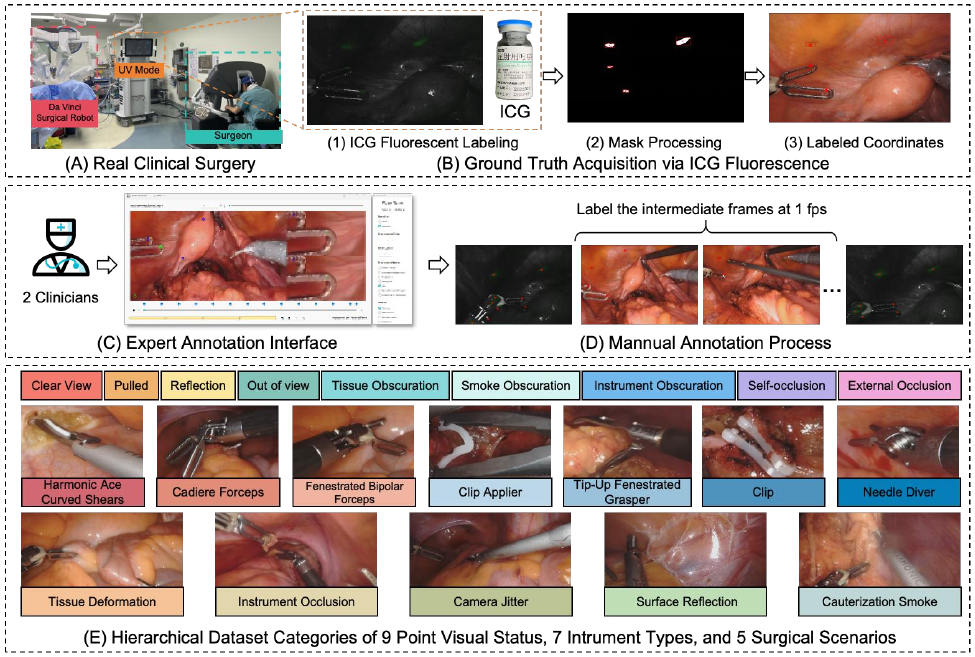}
  \caption{Data collection and annotation workflow for VL-SurgPT. (A) In vivo surgical setup using the da Vinci Xi system. (B) Ground truth acquisition using Indocyanine Green (ICG) fluorescent markers under UV illumination. (C-D) Annotation interface for point tracking and semantic labeling at 1 fps. (E) Coverage of 7 types of surgical instruments, 9 distinct visual status descriptions, and 5 representative challenging scenarios across our dataset.}
    \label{fig:dataset}
\end{figure*}

\subsection{Vision-Language in Surgical Applications}
Vision-Language (VL) integration has emerged as a powerful paradigm in surgical computer vision~\cite{min2025innovating}, advancing video pretraining~\cite{yuan2024hecvl,yuan2024procedure}, scene captioning~\cite{xu2021learning,xu2022rethinking}, and visual question answering~\cite{seenivasan2022surgical,seenivasan2023surgicalgpt,bai2025surgical}. However, this integration remains largely unexplored for low-level tasks like point tracking. Current surgical VL~\cite{zeng2025surgvlm} approaches typically interpret visible elements rather than leveraging language to enhance context awareness under challenging visual conditions.

Our work, \textbf{VL-SurgPT}, addresses this gap by systematically integrating vision and language for surgical point tracking. By providing synchronized visual coordinates with textual status descriptions and developing text-guided tracking methodology, we establish a foundation for semantically-aware tracking systems that overcome purely visual limitations. This opens new avenues for developing more reliable and interpretable tools for computer-assisted interventions.

\section{VL-SurgPT Dataset}\label{sec:dataset}
The VL-SurgPT dataset was meticulously constructed from real-world surgical procedures, involving a multi-stage process of data acquisition, processing, and comprehensive multimodal annotation, as illustrated in Fig.~\ref{fig:dataset}.

\subsection{Data Collection}
The VL-SurgPT dataset originates from \textit{in vivo} robotic surgical procedures conducted at the Department of Gastrointestinal Surgery of the Shenzhen People's Hospital\footnote{All data collection protocols were approved by No. LL-KY-2023121-01 with appropriate patient consent and privacy protection.}.
All procedures, illustrated in Fig.~\ref{fig:dataset} (A), were performed using the da Vinci Xi Surgical System. The source footage encompasses a range of complex operations, including radical gastrectomy, radical resection of gastrointestinal tumors, radical resection of rectal cancer, and radical colectomy. 
Initially, data from 20 distinct surgeries, totaling approximately 33 hours of video, were collected. This raw footage underwent a rigorous quality screening process, prioritizing video clarity, minimal visual artifacts, and procedural consistency relevant to point tracking challenges. Following this curation, the final dataset comprises 115 minutes of high-quality surgical video selected for detailed annotation.

\subsection{Data Processing and Annotation}

The creation of VL-SurgPT involved a detailed two-stage data processing and annotation pipeline, combining intraoperative Indocyanine Green (ICG) marking with postoperative expert manual labeling, as depicted in Fig.~\ref{fig:dataset} (B)-(D).

\textbf{Intraoperative ICG Marking and Ground Truth Acquisition.}
To establish reliable ground truth for point locations, ICG dye, visible under near-infrared (NIR) or UV fluorescence imaging, was utilized.
\textit{For tissue points:} During surgery, a clinician applied ICG dye to 2-4 discrete locations on the tissue surface using an ICG-tipped needle holder.
\textit{For instrument keypoints:} Before surgical use, ICG dye was applied to predefined keypoints (e.g., tip, joints, shaft) on seven types of surgical instruments, typically marking 2–7 keypoints per instrument.

Recognizing the dynamic nature of \textit{in vivo} surgery, which precludes the controlled repetitions possible in \textit{ex vivo} setups (e.g., SurgPose~\cite{wu2025surgpose}), we adopted a strategy to capture ground truth at the beginning and end of short video clips. The protocol involved (see Fig.~\ref{fig:dataset} (B)): (1) Activating the da Vinci system's UV/fluorescence mode with instruments and tissues held static, allowing for clear recording of ICG-marked keypoint coordinates. (2) Switching to standard white light and performing normal surgical maneuvers for 5–10 seconds. (3) Re-activating UV/fluorescence mode with static positioning to record the endpoint ICG coordinates. These ICG-derived coordinates from the first and last frames of each short clip, i.e., the \textbf{query points} and the \textbf{end points}, serve as high-fidelity ground truth anchors. Fluorescent markers were extracted by isolating activated ICG regions via binary thresholding under UV illumination.

\textbf{Postoperative Manual Annotation and Semantic Labeling.}
Following \textit{in vivo} data acquisition, the collected video clips were manually annotated by two experienced clinical doctors using a custom-designed labeling tool (Fig.~\ref{fig:dataset} (C)-(D)). For each clip:
1.  The ICG-marked points in the first frame served as initial query points for tracking.
2.  Clinicians manually tracked these points frame-by-frame through the white-light portion of the clip, annotating their 2D coordinates at 1 frame per second (fps). The ICG-marked points in the final frame provided a reference to validate the trajectory endpoint.
3.  Crucially, alongside coordinate data, each annotated point was assigned multiple semantic labels:
    \begin{itemize}
        \item A \textbf{point type} label (tissue or instrument).
        \item A \textbf{point status} textual description (e.g., ``Clear View'', ``Smoke Obscuration'', ``Pulled'', ``Reflection'', etc., from a predefined vocabulary shown in Fig.~\ref{fig:dataset} (E)).
        \item For instrument points, additional labels included an \textbf{instance identifier} and the \textbf{specific instrument type}.
    \end{itemize}

\subsection{Dataset Details}


VL-SurgPT provides a substantial collection of \textit{in vivo} surgical tracking data spanning diverse clinical scenarios, as summarized in Fig.~\ref{fig:teaser}. Illustrative examples of these scenarios and instruments are visualized in Fig.~\ref{fig:dataset} (E). 

\textbf{Tissue Tracking Subset:} This subset consists of 754 \textit{in vivo} video clips, accumulating to 180.5k frames. From these, 7,117 frames were manually annotated at 1 fps, yielding 1,862 distinct point trajectories of the query points and a total of 17,171 visible tissue points across all the frames of this subset. For tissue tracking, VL-SurgPT encompasses five challenging \textit{in vivo} surgical scenarios: \textit{Tissue Deformation}, \textit{Instrument Occlusion}, \textit{Camera Jitter}, \textit{Surface Reflection}, and \textit{Cauterization Smoke}. Each scenario is well-represented, typically with over 120 video clips, more than 1,200 annotated frames, and between 300 to 450 tracked point trajectories. This balanced distribution across varied intraoperative conditions is designed to facilitate robust evaluation of point tracking models.

\textbf{Instrument Tracking Subset:} This subset includes 154 \textit{in vivo} video clips, totaling 26,490 frames. Manual annotations were performed on 1,108 of these frames. The dataset features 7 distinct types of surgical instruments, with up to 7 predefined keypoints annotated per visible instrument instance in a labeled frame.
For instrument tracking, seven commonly used surgical instruments are annotated: Harmonic Ace Curved Shears, Cadiere Forceps, Fenestrated Bipolar Forceps, Clip Applier, Clip, Tip-Up Fenestrated Grasper, and Needle Driver. Instruments like the Harmonic Ace Curved Shears and Fenestrated Bipolar Forceps have extensive coverage (hundreds of labeled frames and thousands of keypoints each), while less frequently appearing but procedurally critical tools such as the Clip Applier and Needle Driver are also included to ensure comprehensive representation of tool diversity.

\subsubsection{Context-aware Textual Point Label}

Unlike traditional datasets that focus solely on geometric coordinates, VL-SurgPT associates each tracked point with descriptive textual labels, providing crucial semantic context, as demonstrated in Fig.~\ref{fig:teaser}. Specifically, each annotated point includes:
\begin{itemize}
    \item \textbf{2D Coordinates}: Precise pixel location $(x, y)$ in the frame or ``null'' in case the point is invisible.
    \item \textbf{Point Type}: Categorical label (``Tissue'' or ``Instrument'').
    \item \textbf{Point Status}: A textual description from a predefined vocabulary indicating the point's current visual condition (e.g., ``Clear View'', ``Pulled'', ``Reflection'', ``Smoke Obscuration'', ``Instrument Obscuration'', ``Tissue Obscuration'', ``Out of View'', ``External Occlusion'', ``Self-occlusion'').
    \item \textbf{Instrument-Specific Labels}: \textbf{Instrument Type} (e.g., ``Clip Applier'', ``Cadiere Forceps'', etc.) and \textbf{Instance ID} used to distinguish instruments within the same frame.
\end{itemize}

The distribution of semantic labels across different visual conditions reveals important patterns in the challenges present during surgical procedures. For tissue points, while ``Clear View'' and ``Pulled'' statuses predominate, a substantial portion experiences obscuration from instruments and surrounding tissue. Specular reflections from moist tissue surfaces also appear frequently, representing a significant tracking challenge unique to surgical environments. 
For instrument points, the majority appear in clear view—a natural consequence of surgical protocols that prioritize instrument visibility for safe operation. Nevertheless, approximately one-third of annotated instrument points exhibit challenging visual conditions, including external occlusion by tissue or smoke, self-occlusion between instrument parts, and instances where instruments move partially or completely out of the field of view. This significant proportion of visually compromised points underscores the complexity of tool point tracking in dynamic \textit{in vivo} procedures. 

\begin{figure*}[ht]
  \centering
  \includegraphics[width=0.9\linewidth]{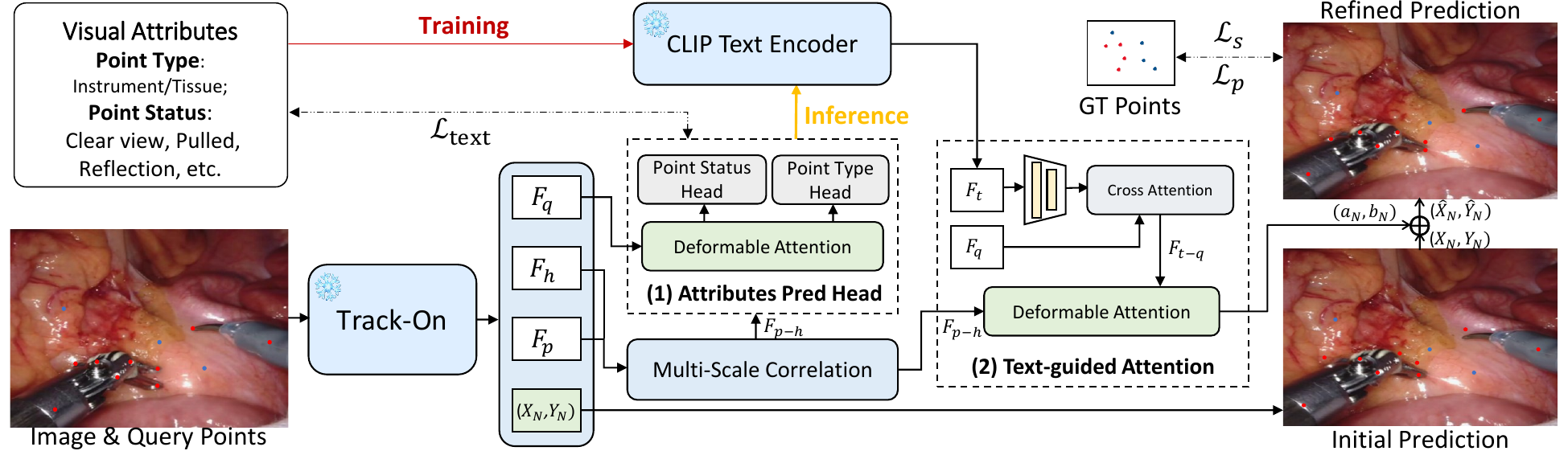}
  \caption{Overview of our Text-guided Surgical Point Tracking (TG-SurgPT). The method builds upon Track-On~\cite{Aydemir2025trackon} by integrating visual features with semantic text descriptions through cross-modal attention.}
\label{fig:tg_method}
\end{figure*}

\section{Multimodal Tracking Baseline: TG-SurgPT}

To address the lack of text-guided reasoning and underutilization of sparse semantic labels in existing point tracking models, we propose Text-Guided Surgical Point Tracking (TG-SurgPT). Our approach builds upon the Track-On~\cite{Aydemir2025trackon} framework, chosen for its balance of efficiency and performance, and extends it to enable text-guided tracking of arbitrary points. 

As illustrated in Fig.~\ref{fig:tg_method}, TG-SurgPT systematically integrates visual and textual modalities through a dual-branch architecture with cross-modal attention mechanisms. \textbf{Textual Branch:} For each tracked point, we encode two types of semantic attributes: \textbf{Point Type} (tissue/instrument) and \textbf{Point Status} (e.g., ``Clear View'', ``Smoke Obscuration'', ``Pulled''). These descriptions are processed through a frozen CLIP Text Encoder~\cite{radford2021learning} to produce textual feature embeddings $F_{t} \in \mathbb{R}^{2 \times 512}$.
\textbf{Visual Branch:} The query points and video frames are processed by the frozen Track-On model, yielding: (1) feature representations of the query points ($F_{q}$), (2) dense visual features of the current frame ($F_{h}$), (3) coarse patch-level matching positions for each query point ($F_{p}$), and (4) initial predicted coordinates $(X_{N}, Y_{N})$, where $N$ is the number of query points.

\subsubsection{Attributes Pred Head}
To enable autonomous semantic understanding during inference, we introduce an Attributes Prediction Head that predicts point status from visual features alone.
As illustrated in Fig.~\ref{fig:tg_method} (1), we first fuse $F_{p}$ and $F_{h}$ using a multi-scale correlation module. The resulting fused representation $F_{p\text{-}h}$ serves as both Key and Value in a Multi-Scale Deformable Attention module, with query features $F_{q}$ as the Query.
This attention mechanism allows each query to attend to a small set of spatially offset sampling points across multiple feature levels. 
We further introduce two parallel classification heads: a \textit{Point Type Head}, which predicts a $2 \times N$ vector indicating the spatial location of each point; and a \textit{Point Status Head}, which outputs a $7 \times N$ or $4 \times N$ matrix depending on the point type (tissue or instrument) for point-wise status classification. 

\subsubsection{Text-guided Attention} 
We design a guided attention module for text-vision feature fusion, as illustrated in Fig.~\ref{fig:tg_method} (2). 
We first project the textual features $F_{t}$ to match the dimensionality of visual query features $F_{q}$ using learned linear transformations. 
We then perform standard cross-attention with $F_{q}$ as the Query and $F_{t}$ as the Key and Value, producing a fused representation $F_{t\text{-}q}$. 
This text-enhanced features $F_{t\text{-}q}$ are combined with the visual-spatial features $F_{p\text{-}h}$ through another deformable attention layer to produce refinement offsets $(a_{N}, b_{N})$.
The final predicted coordinates are computed as:
\begin{equation}
(\hat{X}_{N}, \hat{Y}_{N}) = (X_{N} + a_{N}, Y_{N} + b_{N}).
\end{equation}
It is worth noting that during training, we use ground-truth textual descriptions from the dataset to guide visual understanding. At inference, these descriptions are replaced by the model’s own predicted status labels, enabling text-guided reasoning without relying on manually provided annotations. This design allows the model to improve tracking accuracy in a fully automatic manner during deployment.

\subsubsection{Loss Function} 
Transformer-based models such as Track-On~\cite{Aydemir2025trackon} normally require training on densely annotated synthetic datasets like MOVi-F~\cite{greff2022kubric}, which uses a large number of query points (e.g., 2048 points sampled over 25-frame sequences). 
Our method is capable of training with sparsely labeled real-world surgical data without relying on dense point annotations, benefiting from textual annotations. 
During training, our loss function $\mathcal{L}$ consists of three key components: the point distance loss $\mathcal{L}_{p}$, the trajectory smoothness loss $\mathcal{L}_{s}$, and the textual classification loss $\mathcal{L}_{\text{text}}$. 
The total training loss $\mathcal{L}$ is accumulated across all frames with GT as:
\begin{equation}
\mathcal{L} = 
\sum_{t \in \mathcal{T}}
\big( \underbrace{\mathcal{H}_\delta(\hat{\mathbf{p}}_t - \mathbf{p}_t)}_{\mathcal{L}_{p}} + 
\underbrace{\|\Delta^2 \hat{\mathbf{p}}_t\|_1}_{\mathcal{L}_{s}} +
\underbrace{\mathcal{L}_{\text{CE}}(\hat{\mathbf{s}}_t, \mathbf{s}_t)}_{\mathcal{L}_{\text{text}}} \big),
\end{equation}
where $\mathcal{T}$ denotes the set of selected frames for supervision, $\hat{\mathbf{p}}_t$ and $\mathbf{p}_t$ are the predicted and ground-truth point coordinates, and $\hat{\mathbf{s}}_t$, $\mathbf{s}_t$ denote the predicted and target status labels.
Here, $\mathcal{L}_{p}$ measures the point-wise distance using the robust Huber loss~\cite{meyer2021alternative} $\mathcal{H}_\delta$. $\mathcal{L}_s$ promotes temporal smoothness by minimizing the second-order difference $\Delta^2 \hat{\mathbf{p}}_t$ of predicted trajectories. Finally, $\mathcal{L}_{\text{text}}$ applies cross-entropy loss $\mathcal{L}_{\text{CE}}$ between $\hat{\mathbf{s}}_t$ and $\mathbf{s}_t$ to supervise semantic status classification, such as ``Occluded'' or ``Off-camera''.

\begin{table*}[t]
\centering
\setlength{\tabcolsep}{1mm}
\begin{tabular}{lccccccccc}
\hline
\multirow{2}{*}{\textbf{Method}} & \multicolumn{4}{c}{\textbf{Tissue Subset}} & \multicolumn{4}{c}{\textbf{Instrument Subset}} & \multirow{2}{*}{\textbf{Mean fps}} \\
\cmidrule(lr){2-5}
\cmidrule(lr){6-9}
           & \textbf{AJ $\uparrow$} & \textbf{$<\delta^{x}_{avg}$ $\uparrow$} & \textbf{OA $\uparrow$} & \textbf{EPE $\downarrow$} 
                & \textbf{AJ $\uparrow$} & \textbf{$<\delta^{x}_{avg}$ $\uparrow$} & \textbf{OA $\uparrow$} & \textbf{EPE $\downarrow$} &  \\
\hline
RAFT~\cite{teed2020raft}                    & 27.81             & 30.37            & 85.81             & 99.73             & 23.67             & 33.20            & 76.49 & 138.37 & 12.47 \\
SEA-RAFT~\cite{wang2024sea}                 & 20.49             & 24.08            & 83.14             & 89.52             & 18.11             & 27.07            & 78.02 & 103.18 & 13.25 \\
TAPIR~\cite{doersch2023tapir}               & 40.01             & 46.63            & 70.63             & 56.99             & 41.81             & 49.49            & 76.15 & 75.78  & \underline{13.41} \\
BootsTAP~\cite{doersch2024bootstap}         & 56.93             & 62.77            & 87.87             & 23.52             & 46.26             & 53.71            & 84.76 & 44.20  & \textbf{14.25} \\
CotrackerV3~\cite{karaev2024cotracker3}     & 43.27             & 38.11            & 76.52             & 56.13             & 42.03             & 38.31            & 79.23 & 52.12  & 9.87  \\
MFT~\cite{neoral2024mft}                    & 57.61             & \underline{67.12}   & \underline{90.46}    & 18.07 & 45.63             & 55.18            & \underline{86.13} & \underline{40.97}  & 1.42  \\
MFTIQ~\cite{serych2025mftiq}                & \underline{61.52}    & 63.44            & 87.80             & 19.81             & 46.56             & 56.47 & 74.21 & \underline{39.42}  & 0.71  \\
Track-On~\cite{Aydemir2025trackon}& 58.55 & 66.27 & 88.81 & \underline{13.79}    & \underline{46.97}    & \underline{59.18} & 85.07 & 41.67  & 10.85 \\
\hline
TG-SurgPT (Ours) & \textbf{62.88} & \textbf{67.77} & \textbf{91.04} & \textbf{11.02}  & \textbf{49.52} & \textbf{62.94}  & \textbf{89.79} & \textbf{39.14} & 9.72\\
\hline
\end{tabular}
\caption{Benchmark of tracking performance on tissue and instrument points. Our multimodal (Vision-Language) TG-SurgPT shows superiority over unimodal (vision-only) baselines, with a decent inference speed.}
\label{tab:benchmark}
\end{table*}

\section{Experiments}\label{sec:experiments}

\subsection{Experimental Setup}
We conduct comprehensive experiments to establish baseline performance on VL-SurgPT and demonstrate the effectiveness of text-guided tracking. Our evaluation encompasses eight state-of-the-art vision-based tracking methods (RAFT~\cite{teed2020raft}, SEARAFT~\cite{wang2024sea}, MFT~\cite{neoral2024mft}, MFTIQ~\cite{serych2025mftiq}, TAPIR~\cite{doersch2023tapir}, BootsTAP~\cite{doersch2024bootstap}, CoTrackerV3~\cite{karaev2024cotracker3}, and Track-On~\cite{Aydemir2025trackon}) and our proposed \textbf{TG-SurgPT}, a novel text-guided approach that leverages semantic descriptions.
We follow standard metrics from point tracking literature (TAP-Vid~\cite{doersch2022tapvid} and STIR~\cite{schmidt2024surgical}) for reliable evaluation, including \textit{Average Position Accuracy} ($< \delta_{\text{avg}}^x$), \textit{Average Jaccard} (AJ), \textit{Occlusion Accuracy}, \textit{End Point Error} (EPE), \textit{Inference Speed} (fps). 

The dataset is split 4:1 for training and testing, stratified by surgical scenarios (tissue subset) and instrument types (instrument subset) to promote balanced distribution.
All experiments are conducted on a single NVIDIA RTX 4090 GPU with the PyTorch framework. All input frames are resized to $720 \times 480$ resolution for consistent evaluation. We use default parameters recommended by the original authors for all baseline methods to ensure fair comparison, with consistent input preprocessing and evaluation protocols.

\begin{figure}[t]
  \centering
  \includegraphics[width=\linewidth]{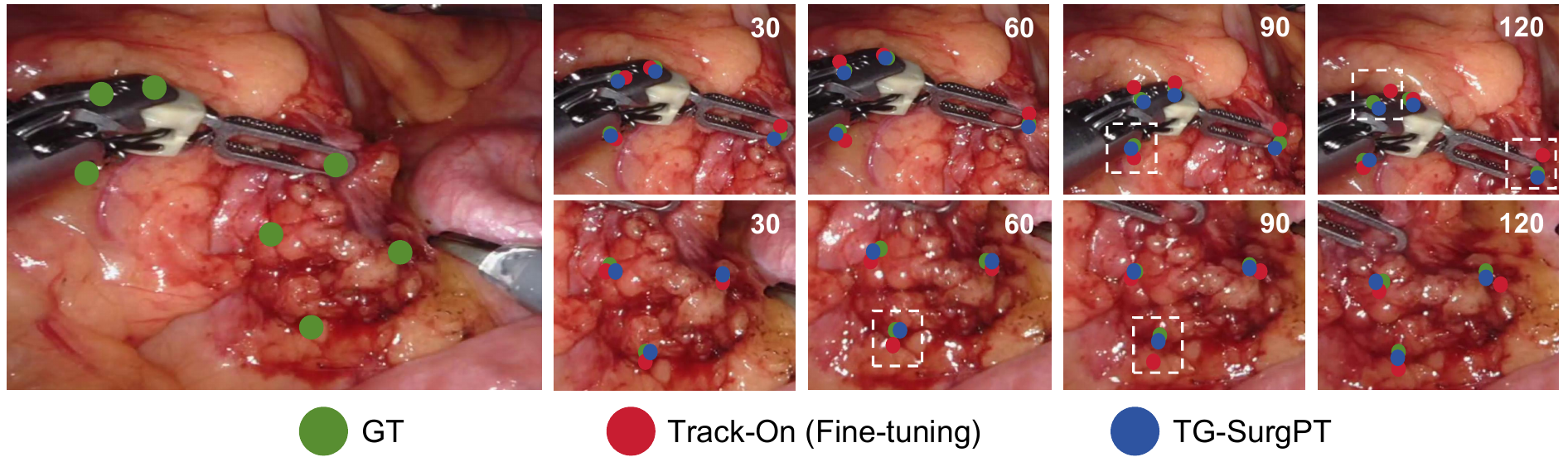}
  \caption{Qualitative comparison of tracking performance of the finetuned Track-On~\cite{Aydemir2025trackon} and our TG-SurgPT across sequential frames (30, 60, 90, 120) for both instrument (top row) and tissue (bottom row) point tracking.}
\label{fig:quality_res}
\end{figure}

\begin{figure}[t]
  \centering
  \includegraphics[width=\linewidth]{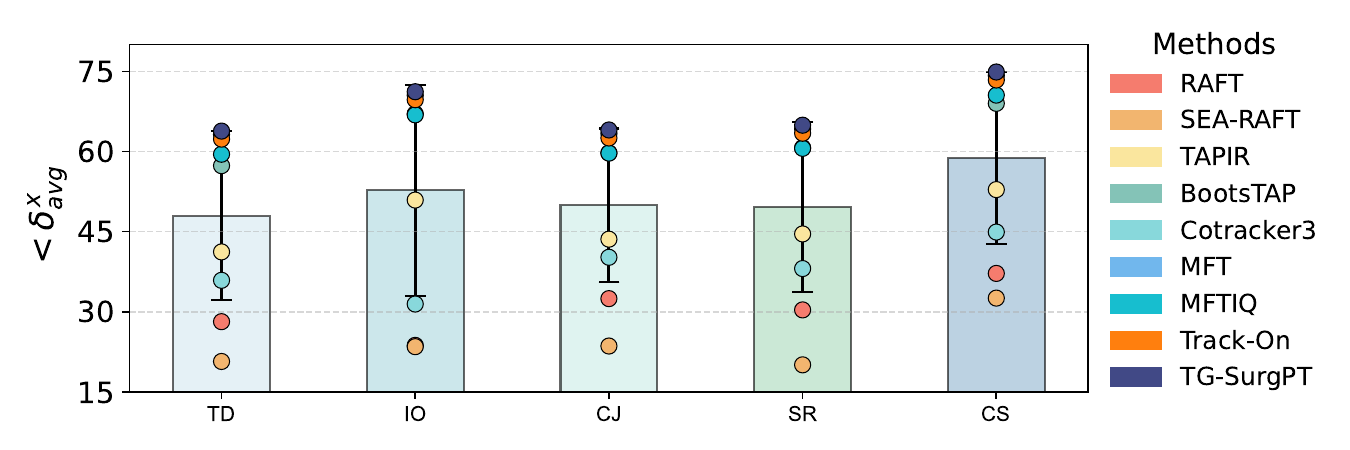}
  \caption{Scenario-specific comparison ($<\delta^{x}_{avg}$, higher is better) across Tissue Deformation (TD), Instrument Occlusion (IO), Camera Jitter (CJ), Surface Reflection (SR), and Cauterization Smoke (CS). Box height indicates the mean performance of all methods.}
\label{fig:scence_compare_fig}
\end{figure}

\subsection{Results and Analysis}
\subsubsection{Benchmark Comparison}
Table~\ref{tab:benchmark} presents a detailed performance comparison across eight state-of-the-art tracking methods and our proposed TG-SurgPT approach. 
The results reveal several important patterns in surgical point tracking performance.
Traditional flow-based methods (RAFT, SEA-RAFT) demonstrate substantial limitations in surgical environments, with Average Jaccard scores consistently below 30\%, highlighting their inability to handle complex tissue deformations and visual artifacts inherent to surgical procedures.
In contrast, transformer-based approaches show markedly superior performance, underscoring the effectiveness of attention mechanisms in capturing the complex spatial-temporal relationships in surgical scenes. For tissue tracking, MFTIQ achieves the highest baseline AJ (61.52\%), while MFT excels in position accuracy (67.12\%) and occlusion detection (90.46\%). Track-On demonstrates optimal trajectory completion with the lowest EPE (13.79 px). Similar patterns appear in instrument tracking, where Track-On leads in spatial accuracy metrics while MFT maintains superior occlusion handling capabilities. However, a critical efficiency-accuracy trade-off emerges: while MFT and MFTIQ achieve strong accuracy, their impractical inference speeds (1.42-0.71 fps) preclude real-time clinical deployment. Track-On provides the optimal balance at 10.85 fps, making it suitable for surgical integration.

Our multimodal TG-SurgPT consistently outperforms all baselines across every metric while maintaining practical processing speed (9.72 fps). The improvements are particularly pronounced for instrument tracking (5.4\% AJ gain, 6.4\% position accuracy improvement), demonstrating that semantic text guidance provides contextual information unavailable to purely visual methods. These results validate our core hypothesis that incorporating point status descriptions significantly enhances tracking robustness under challenging surgical conditions where conventional vision-only approaches fail.
Figure~\ref{fig:quality_res} compares TG-SurgPT with fine-tuned Track-On across sequential frames, demonstrating TG-SurgPT's enhanced trajectory coherence and accuracy in tracking instruments and tissue points.

\subsubsection{Scenario-specific Results Analysis}

Figure~\ref{fig:scence_compare_fig} reveals the relative difficulty hierarchy across five challenging surgical conditions. Tissue Deformation emerges as the most challenging scenario, where complex non-rigid transformations fundamentally challenge current visual tracking paradigms. Surprisingly, Cauterization Smoke (CS) yields the highest performance (58.62\% mean) despite intuitive expectations of visual impairment, likely because smoke creates distinctive temporal patterns and edge features that attention mechanisms can effectively leverage. 
The consistent performance gaps between methods across scenarios highlight fundamental algorithmic differences. Flow-based methods show severe degradation in all conditions, while transformer-based approaches exhibit scenario-dependent variations. TG-SurgPT demonstrates consistent improvements across all scenarios, particularly in dynamic conditions like IO and CJ. This suggests that semantic descriptions provide crucial disambiguation when visual features become unreliable. The results strongly support our hypothesis that text guidance is most beneficial in situations where visual-only methods face the greatest challenges.

\begin{figure}[t]
  \centering
  \includegraphics[width=\linewidth]{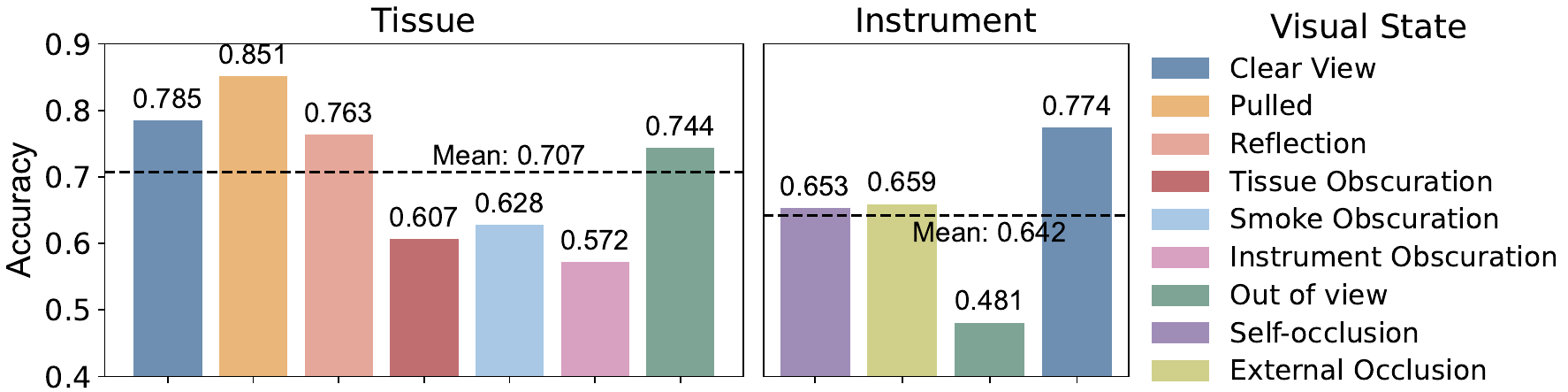}
  \caption{Classification accuracy of different visual status categories for both tissue and instrument subsets.}
\label{fig:text_classification_res}
\end{figure}

\subsubsection{Visual Status Prediction Analysis}
Figure~\ref{fig:text_classification_res} reports the prediction accuracy for each visual status category across the tissue and instrument subsets with our TG-SurgPT. For tissue points (left), distinctive states like \textit{Pulled} (85.1\%) and \textit{Clear View} (78.5\%) achieve high accuracy, while ambiguous obscuration states show lower performance, suggesting difficulty in disambiguating between different types of visual obstruction when tissue points become partially or completely hidden. For instrument points (right), both \textit{Self-occlusion} (65.3\%) and \textit{External Occlusion} (65.9\%) achieve decent performance, showing the model's ability to distinguish between different types of occlusion scenarios. However, \textit{Out of View} exhibits the poorest accuracy (48.1\%), highlighting the challenge of correctly identifying when instrument points move outside the field of view, particularly during rapid instrument movements or camera transitions.
Overall, while our TG-SurgPT model demonstrates promising semantic understanding capabilities across most visual status categories, the results reveal distinct challenges for each point type: tissue points struggle most with occlusion disambiguation, while instrument points face difficulties in out-of-view detection. These findings indicate specific areas for improvement in fine-grained visual status classification.

\subsubsection{Ablation Study}

Table~\ref{tab:ablation_text_guidance} summarizes ablation studies on the contributions of fine-tuning, training clip length, and text guidance in TG-SurgPT for tissue and instrument tracking. \textbf{Fine-tuning Impact:} Fine-tuning consistently improves performance across all metrics. For tissue tracking, short-clip fine-tuning increases AJ from 58.55 to 61.09 (+4.3\%) and reduces EPE from 13.79 to 12.82 px. Similar gains are observed for instrument tracking (AJ: 46.97$\rightarrow$47.48, +1.1\%). \textbf{Clip Length Analysis:} Short clips (31 frames per clip) generally outperform long clips (181 frames per clip) in most metrics, particularly for AJ and EPE. This suggests that shorter temporal windows reduce noise and improve point correspondence stability, making them more suitable for sparse surgical annotations. \textbf{Text Guidance Effectiveness:} Text guidance provides substantial improvements across both subsets. For tissue tracking, adding text guidance to short-clip fine-tuning yields the best performance: AJ increases to 62.88 (+2.9\% over visual-only), OA reaches 91.04 (+1.3\%), and EPE drops to 11.02 px (-14.0\%). For instrument tracking, text guidance achieves AJ of 49.52 (+4.3\%) and OA of 89.79 (+3.9\%), demonstrating that semantic context significantly enhances tracking robustness.

\begin{table}[t]
\centering
\setlength{\tabcolsep}{0.65mm}
\begin{tabular}{ccccccc}
\toprule
\textbf{Finetuning} & \textbf{Clip Size} & \textbf{Text} 
& \textbf{AJ $\uparrow$} & \textbf{$<\delta^{x}_{avg}$ $\uparrow$} & \textbf{OA $\uparrow$} & \textbf{EPE $\downarrow$} \\
\hline
\multicolumn{7}{l}{Tissue Subset} \\
\hline
$\times$     & $\times$     & $\times$     & 58.55 & 66.27 & 88.81 & 13.79 \\
$\checkmark$ & short & $\times$     & 61.09 & 66.39 & 89.70 & 12.82 \\
$\checkmark$ & long  & $\times$     & 59.91 & 66.73 & 89.41 & 12.94 \\
$\checkmark$ & short & $\checkmark$ & \textbf{62.88} & \textbf{67.77} & \textbf{91.04} & \textbf{11.02} \\
$\checkmark$ & long  & $\checkmark$ & \underline{62.09} & \underline{67.46} & \underline{90.88} & \underline{11.77} \\
\hline
\multicolumn{7}{l}{Instrument Subset} \\
\hline
$\times$     & $\times$   & $\times$     & 46.97 & 59.18  & 85.07 & 41.67 \\
$\checkmark$ & short & $\times$     & 47.48 & 60.75  & 86.43 & 40.41 \\
$\checkmark$ & long  & $\times$     & 47.22 & 59.89  & 87.01 & 40.82 \\
$\checkmark$ & short & $\checkmark$ & \textbf{49.52} & \underline{62.94}  & \textbf{89.79} & \textbf{39.14} \\
$\checkmark$ & long  & $\checkmark$ & \underline{48.77} & \textbf{63.08}  & \underline{88.75} & \underline{39.69} \\
\hline
\end{tabular}
\caption{Ablation study on fine-tuning, clip length, and text guidance for tissue and instrument subsets.}
\label{tab:ablation_text_guidance}
\end{table}

\section{Discussion and Conclusion}

We present \textbf{VL-SurgPT}, the first large-scale multimodal dataset for surgical point tracking that systematically integrates visual trajectories with semantic point status descriptions. Through comprehensive benchmarking and the proposed text-guided tracking method \textbf{TG-SurgPT}, we demonstrate the significant potential of incorporating semantic understanding into surgical scene analysis, paving the way for more robust and context-aware point tracking in computer-assisted surgery.
Current limitations include the dataset's limited scope (primarily gastrointestinal procedures using the da Vinci system) and TG-SurgPT's inference speed of 9.72 fps, which approaches but does not meet real-time clinical requirements. Future work will expand the dataset to additional surgical domains, optimize inference speed for real-time clinical use, and develop more advanced cross-modal learning strategies to better leverage limited labeled data.

\section{Acknowledgments} 
This work was supported in part by Hong Kong Research Grants Council (RGC) Collaborative Research Fund (CRF C4026-21GF), Research Impact Fund RIF R4020-22 and General Research Fund (GRF 14216022, 14204524, 14203323, 14206125), NSFC Young Scientists Fund - Category A T252500134, NSFC/RGC Joint Research Scheme N\_CUHK420/22; Guangdong Basic and Applied Basic Research Foundation (GBABF) \#2021B1515120035.

\bibliography{aaai2026}




\clearpage
\twocolumn[
\begin{center}
{\LARGE \bf Supplementary Material \par}
\vspace{1em}
\end{center}]

\begin{table*}[!ht]
\centering
\resizebox{\linewidth}{!}{
\begin{tabular}{l|ccccccccc}
\toprule
\multicolumn{10}{c}{\textbf{Tissue}} \\
\midrule
\textbf{Dataset} & \textbf{Recording Condition} & \textbf{Annotation Level} & \textbf{Marker} & \textbf{Number of Videos} & \textbf{Video Frames} & \textbf{Label Frames} & \textbf{Point Tracks} & \textbf{Total Points} & \textbf{FPS}  \\
\midrule
SuPer~\cite{li2020super}         & \textit{ex vivo}   & V   & SW    & 1                     & 522                       & 52                & 20                & 1024              & 1  \\
Sem. SuPer~\cite{lin2023semantic}    & \textit{ex vivo}  & V    & Beads    & 4                     & 600                       & 600               & 240               & 6750              & 1  \\
SurgT~\cite{cartucho2024surgt}         & \textit{in vivo}   & V  & SW    & 32                    & 24.7k                     & \textbf{21176}    & 32                & \underline{21176}    & 30  \\
STIR~\cite{schmidt2024surgical}   & \textit{in/ex vivo}  & V  & IR    & \underline{136 / 436} & \textbf{262.5k/ 63.15k}   & 576               & \textbf{3604}     & 7208              & FL  \\
SurgMotion~\cite{zhan2024surgmotion} & \textit{in vivo} & V  & SW  & 20  & 2440  & 2440  & 300 & \textbf{36600}  & 30\\
VL-SurgPT(Ours)           & \textit{in vivo}   & VL   & IR   & \textbf{754}          & \underline{180.5k}        & \underline{7117}  & \underline{1862}  & 17171 & 1  \\
\midrule
\multicolumn{10}{c}{\textbf{Surgical Instruments}} \\
\midrule
\textbf{Dataset} & \textbf{Recording Condition} & \textbf{Annotation Level} & \textbf{Marker} & \textbf{Number of Videos} & \textbf{Video Frames} & \textbf{Label Frames} & \textbf{Keypoints} & \textbf{Types of Instruments} & \textbf{FPS} \\
\midrule
EndoVis15~\cite{bodenstedt2018comparative}        & \textit{ex vivo}  & V    & SW   & 6               & 9000              & 360               & 1             & 2          & 1  \\
EndoVis15~\cite{bodenstedt2018comparative}        & \textit{in vivo}  & V    & SW   & 6               & 360               & 60                & 1             & 2          & 1  \\
Du et al. 2018~\cite{du2018articulated}   & \textit{ex vivo}  & V    & SW   & 8               & 9000              & 1850  & 5             & 2          & 1  \\
SurgPose~\cite{wu2025surgpose}         & \textit{ex vivo}   & V   & UV   & \underline{33}  & \textbf{60000}    & \textbf{60000}    & \textbf{7}    & \underline{6} & 30  \\
SurgMotion~\cite{zhan2024surgmotion} & \textit{in vivo}  & V  & SW  & 20  & 2440  & 2440  & 5 & —  & 30   \\
VL-SurgPT(Ours)             & \textit{in vivo}   & VL   & IR  & \textbf{154}    & \underline{26490} & \underline{1108}         & \textbf{7}    & \textbf{7} & 1  \\
\bottomrule
\end{tabular}
}
\caption{Comprehensive comparison of public surgical point tracking datasets. VL-SurgPT is the first dataset to provide synchronized vision-language annotations for both tissue and instrument tracking in \textit{in vivo} conditions. \textit{Recording Condition:} ex vivo = controlled laboratory environment, in vivo = live surgical procedures. \textit{Marker Type:} SW = manual software annotation, Beads = physical 2mm steel beads, IR = infrared contrast dye, UV = ultraviolet fluorescent dye. \textit{Annotation Level:} V = vision-only coordinate annotations, VL = multimodal vision-language with semantic descriptions.}
\label{tab:dataset_comparison}
\end{table*}

\section{Dataset Details}

\subsection{Comparison of Existing Datasets}

Existing surgical point tracking datasets can be broadly categorized into two types: tissue point tracking and instrument keypoint tracking, as summarized in Table~\ref{tab:dataset_comparison}.

For \textbf{tissue point tracking}, early datasets such as SuPer~\cite{li2020super} were constructed under controlled \textit{ex vivo} conditions, with sparse manual annotations. Semantic SuPer~\cite{lin2023semantic} introduced grid-pattern markers to enhance annotation precision, yet it remains limited to \textit{ex vivo} scenarios and short sequences. SurgT~\cite{cartucho2024surgt} contributed a larger-scale \textit{in vivo} dataset but primarily annotated bounding boxes rather than continuous tissue point trajectories. STIR~\cite{schmidt2024surgical} utilized infrared markers to improve point localization in both \textit{in vivo} and \textit{ex vivo} settings, yet the annotations are mostly sparse and located at sequence endpoints, lacking dense temporal coverage and semantic condition labels. SurgMotion~\cite{zhan2024surgmotion} provides high-density point tracks in short \textit{in vivo} videos, but it does not offer semantic descriptions of the visual scene.

Our proposed VL-SurgPT dataset addresses these limitations by offering a \textbf{large-scale}, \textbf{vision-language} benchmark focused on \textit{\textbf{in vivo}} tissue point tracking. It contains over 180,000 video frames from 754 procedures, with 1,862 annotated tracks across 7,117 frames. In addition to spatial trajectories, each annotated point is accompanied by semantic descriptors that characterize its visual status, such as occlusion, smoke, deformation, or reflection. This enables detailed analysis of tracker performance under various real-world surgical conditions.

For \textbf{instrument keypoint tracking}, datasets such as EndoVis15~\cite{bodenstedt2018comparative} and Du et al.~\cite{du2018articulated} offer annotated keypoints in \textit{ex vivo} setups, with limited semantic context. SurgPose~\cite{wu2025surgpose} introduces efficient labeling methods and expands the diversity of instrument types, but remains focused on controlled laboratory environments. Most existing datasets lack annotations of challenging visual states during actual procedures, such as partial occlusion, self-occlusion, or out-of-view conditions.

VL-SurgPT fills this gap by providing \textit{in vivo} instrument keypoint annotations with semantic descriptions. It includes seven distinct instruments, each tracked in over 26,000 frames from 154 procedures. The semantic labels cover a variety of difficult conditions, enabling the development of robust, condition-aware keypoint tracking algorithms. Unlike pose estimation datasets that rely on full 6-DoF modeling, VL-SurgPT focuses directly on 2D keypoint-level challenges, which are more relevant for many vision-language reasoning tasks and practical downstream applications.

\subsection{Annotation Scheme}
\begin{figure*}[ht]
  \centering
  \includegraphics[width=\linewidth]{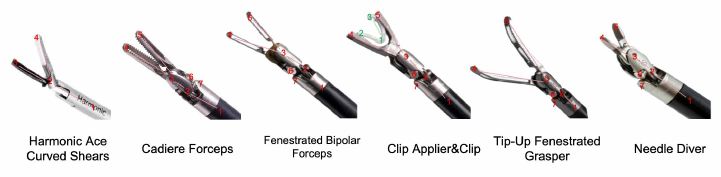}
  \caption{Keypoint annotation scheme for seven types of surgical instruments in VL-SurgPT. Each instrument type has predefined keypoints marking critical anatomical features: single-jointed tools like Harmonic Ace Curved Shears use 4 keypoints, while multi-jointed instruments like Cadiere Forceps employ up to 7 keypoints to capture articulation and pose variations. The annotation follows clinical relevance and kinematic importance for robust instrument tracking.}
\label{fig:annotation_instrument}
\end{figure*}

The annotation scheme distinguishes between two general types of surgical instruments, as illustrated in Figure~\ref{fig:annotation_instrument}. For single-jointed tools like the Harmonic Ace Curved Shears, four keypoints are annotated to capture essential structural features. In contrast, for multi-jointed instruments such as the Cadiere Forceps, five primary keypoints are labeled along with two additional auxiliary points, following the convention in SurgPose~\cite{wu2025surgpose}, to better represent articulation and pose variation. This setup allows for a reliable reconstruction of the instrument's kinematic skeleton and enhances the interpretability of its motion and orientation in dynamic scenes. Additionally, considering the high frequency and functional importance of surgical clips, we also annotate the combination of the clip and its applying tool, the clip applier, as a coherent unit.

For tissue data, annotated points are selected at regions that hold clear clinical relevance and are visually stable throughout the procedure, similar to SurgT~\cite{cartucho2024surgt}. In particular, we focus on anatomically meaningful locations such as the boundaries between vascular structures and darkened tissue regions. These areas are often used by surgeons as visual cues for navigation, dissection, or decision-making, making them ideal reference points for evaluating tracking performance. Their distinct appearance and relative consistency across frames ensure both practical interpretability and high visibility in surgical footage.

In summary, while prior datasets have made important contributions to surgical point tracking, they often lack the scale, annotation granularity, or semantic richness necessary for analyzing tracker behavior in complex \textit{in vivo} environments. VL-SurgPT addresses these issues by providing dense, semantically labeled point trajectories for both tissue and instruments, thereby facilitating more comprehensive and realistic evaluations.

\subsection{Dataset Statistics}
VL-SurgPT provides a substantial collection of \textit{in vivo} surgical tracking data. Table~\ref{tab:dataset_statistics} provides a breakdown of the dataset statistics, categorized by surgical scenarios for tissue tracking and by instrument types for instrument tracking. 

\begin{table}[!ht]
\centering
\setlength{\tabcolsep}{1mm}
\resizebox{\linewidth}{!}{
\begin{tabular}{lcccc}
\toprule
\textbf{} & \textbf{\# Videos} & \textbf{\# Frames} & \textbf{\# Query Points} & \textbf{\# Total Points} \\
\midrule
\multicolumn{5}{l}{\textbf{Surgical Scenarios} (Tissue Tracking Subset)} \\
\midrule
Tissue Deformation           & 139 & 1331 & 352 & 3289 \\
Instrument Occlusion         & 140 & 1320 & 334 & 3084 \\
Camera Jitter                & 166 & 1495 & 427 & 3671 \\
Surface Reflection           & 122 & 1228 & 311 & 3139 \\
Cauterization Smoke          & 187 & 1743 & 438 & 3988 \\
\textbf{Total}      & \textbf{754} & \textbf{7117} & \textbf{1862} & \textbf{17171} \\
\midrule
\multicolumn{5}{l}{\textbf{Instrument Types} (Instrument Tracking Subset)} \\
\midrule
Cadiere Forceps              & 30  & 223  & 87  & 634 \\
Fenestrated Bipolar Forceps  & 75  & 547  & 220 & 1597 \\
Harmonic Ace Curved Shears   & 104 & 753  & 339 & 2427 \\
Tip-Up Fenestrated Grasper   & 23  & 170  & 63  & 478  \\
Clip                         & 42  & 309  & 166 & 1235 \\
Clip Applier                 & 9   & 69   & 15  & 118  \\
Needle Diver                 & 24  & 161  & 139 & 938  \\
\textbf{Total}  & \textbf{307} & \textbf{2232} & \textbf{1029} & \textbf{7427} \\
\bottomrule
\end{tabular}
}
\caption{Dataset statistics of surgical scenarios and instrument types of tissue and instrument tracking subsets.}
\label{tab:dataset_statistics}
\end{table}

\begin{figure*}[!ht]
  \centering
  \includegraphics[width=.9\linewidth]{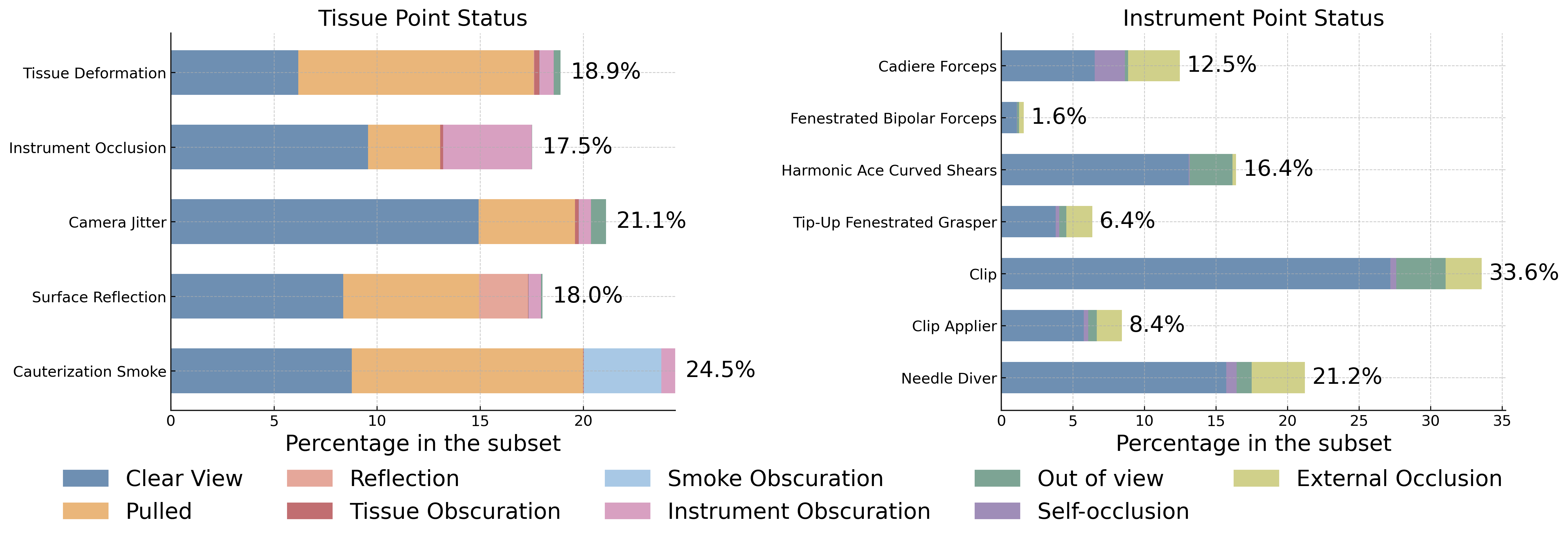}
  \caption{Distribution of semantic point status across surgical scenarios and instrument types in VL-SurgPT.}
\label{fig:status_statistics}
\end{figure*}

\textbf{Tissue Tracking Subset:} This subset consists of 754 \textit{in vivo} video clips, accumulating to 180.5k frames. From these, 7,117 frames were manually annotated at 1 fps, yielding 1,862 distinct point trajectories of the query points and a total of 17,171 visible tissue points across all the frames of this subset. For tissue tracking, VL-SurgPT encompasses five challenging \textit{in vivo} surgical scenarios: \textit{Tissue Deformation}, \textit{Instrument Occlusion}, \textit{Camera Jitter}, \textit{Surface Reflection}, and \textit{Cauterization Smoke}. Each scenario is well-represented, typically with over 120 video clips, more than 1,200 annotated frames, and between 300 to 450 tracked point trajectories. This balanced distribution across varied intraoperative conditions is designed to facilitate robust evaluation of point tracking models.

\textbf{Instrument Tracking Subset:} This subset comprises 307 instrument-wise annotated clips extracted from in vivo surgical videos, covering a total of 2,232 frames with 1,029 query points and 7,427 labeled points. Manual annotations were performed on 1,108 of these frames. The dataset features 7 distinct types of surgical instruments, with up to 7 predefined keypoints annotated per visible instrument instance in a labeled frame.
For instrument tracking, seven commonly used surgical instruments are annotated: Harmonic Ace Curved Shears, Cadiere Forceps, Fenestrated Bipolar Forceps, Clip Applier, Clip, Tip-Up Fenestrated Grasper, and Needle Driver. Instruments like the Harmonic Ace Curved Shears and Fenestrated Bipolar Forceps have extensive coverage (hundreds of labeled frames and thousands of keypoints each), while less frequently appearing but procedurally critical tools such as the Clip Applier and Needle Driver are also included to ensure comprehensive representation of tool diversity.

\subsection{Semantic Point Status Distribution}

The distribution of semantic point statuses, as visualized in Figure~\ref{fig:status_statistics}, illustrates the diverse visual challenges encountered during surgical procedures. 

For \textit{tissue points}, the most frequently observed statuses are ``Clear View'' and ``Pulled'', which together account for a large portion of the annotations. This indicates that, despite ongoing tissue manipulation, the majority of tissue points remain at least partially visible. However, a notable proportion of the points are subject to visual degradation caused by occlusions from surgical instruments or adjacent anatomical structures. In addition, specular reflections on moist tissue surfaces and camera-induced jitter frequently appear, each affecting more than 18\% of the subset. These factors, which are intrinsic to endoscopic imaging, present considerable challenges for robust point tracking. Furthermore, the presence of ``Cauterization Smoke'' in 24.5\% of annotated tissue points reflects the domain-specific difficulties inherent in real-world surgical environments.

For \textit{instrument points}, the dominant status is ``Clear View'', consistent with clinical practices that emphasize maintaining visibility of surgical tools for safe and effective operation. Nevertheless, approximately one-third of instrument points are affected by adverse visual conditions. These include external occlusions caused by tissue or surgical smoke, self-occlusion resulting from complex tool articulation, and cases where the instrument tip partially or fully exits the camera's field of view. Certain instruments, such as the ``Clip'' and ``Needle Driver'', exhibit higher proportions of compromised visibility (33.6\% and 21.2\%, respectively), likely due to their frequent usage in deep or obstructed surgical regions.

In summary, while clear visibility is the most prevalent condition across both categories, a substantial fraction of points are influenced by occlusions and other degradations. This underscores the importance of designing point tracking models that are resilient to the complex, dynamic, and visually ambiguous nature of intraoperative environments.

\section{Experimental Details}
\subsection{Metrics Definition}

To ensure comprehensive and reliable assessment, we adopt standard evaluation metrics from both general computer vision point tracking literature (TAP-Vid~\cite{doersch2022tapvid}) and surgical-specific tracking benchmarks (STIR~\cite{schmidt2024surgical}). Our evaluation protocol encompasses five key metrics that collectively assess spatial accuracy, temporal consistency, occlusion handling, and computational efficiency:

\begin{itemize}

\item \textit{Average Position Accuracy} ($< \delta_{\text{avg}}^x$): Mean percentage of visible points whose predicted positions lie within  2, 4, 8, 16, and 32 pixel distances from the ground truth locations. Following the PCK-style~\cite{yang2012articulated} evaluation, the metric assumes input frames are resized to 256×256 and computes accuracy only over visible points.
\begin{equation}
< \delta_{\text{avg}}^x = \frac{1}{|\mathcal{K}|} \sum_{k \in \mathcal{K}} \left( \frac{1}{|\mathcal{V}|} \sum_{(i,t) \in \mathcal{V}} \mathbb{1} \left( \left\| \hat{\mathbf{p}}_i^{(t)} - \mathbf{p}_i^{(t)} \right\|_2 \leq k \right) \right),
\end{equation}
where $\mathcal{K} = \{2, 4, 8, 16, 32\}$ is the set of distance thresholds (in pixels), $\mathcal{V}$ denotes the set of visible point-frame pairs, and $\|\cdot\|_2$ denotes the Euclidean norm.

\item \textit{Average Jaccard} (AJ): Mean Jaccard score across distance thresholds;
\begin{equation}
\text{AJ} = \frac{1}{|\mathcal{K}|} \sum_{k \in \mathcal{K}} \frac{\text{TP}_k}{\text{TP}_k + \text{FP}_k + \text{FN}_k},
\end{equation}
where $\text{TP}_k$, $\text{FP}_k$, and $\text{FN}_k$ represent true positives, false positives, and false negatives at threshold $k$, respectively.

\item \textit{Occlusion Accuracy} (OA): Accuracy of visibility predictions for all points, including both visible and occluded points;
\begin{equation}
\text{OA} = \frac{1}{|\mathcal{P}|} \sum_{i \in \mathcal{P}} \mathbb{1}[\hat{v}_i = v_i],
\end{equation}
where $\mathcal{P}$ represents all point instances, $\hat{v}_i$ and $v_i$ denote predicted and ground truth visibility states, respectively.

\item \textit{End Point Error} (EPE): Average Euclidean distance between predicted and ground-truth positions of the sequence end points; 
\begin{equation}
\text{EPE} = \frac{1}{|\mathcal{P}|} \sum_{i \in \mathcal{P}} \left\| \hat{\mathbf{x}}_i - \mathbf{x}_i \right\|_2,
\end{equation}
where $\hat{\mathbf{x}}_i$ and $\mathbf{x}_i$ represent predicted and ground truth endpoint coordinates.

\item \textit{Inference Speed} (fps): We measure computational efficiency as the average number of frames processed per second during inference, crucial for real-time surgical applications.

\item \textit{Text Classification Accuracy} ($Acc_{\text{text}}$): For our text-guided method, we additionally evaluate the accuracy of semantic status prediction, measuring how well the model identifies point conditions.


\end{itemize}

\subsection{Implementation Details}
All experiments are conducted on a single NVIDIA RTX 4090 GPU using the PyTorch 2.6 framework with CUDA 12.4. 
Input video frames are uniformly resized to a resolution of $720 \times 480$ to ensure consistency across all methods. We adopt the official implementations and follow the default hyperparameters recommended by the original authors for all baseline methods (e.g., MFT~\cite{neoral2024mft}, Track-On~\cite{Aydemir2025trackon}), unless otherwise stated. All models are evaluated under identical data preprocessing pipelines, including normalization, resizing, and temporal sampling strategies, to eliminate confounding factors.
For fair and reproducible comparison, we apply a unified evaluation protocol across all methods, computing metrics such as $<\delta^x_{k}$ accuracy at multiple thresholds ($k = 2, 4, 8, 16, 32$) and end-point error (EPE) per frame. All results are reported as averages across all annotated points and frames within each scenario. To account for randomness in initialization or inference, all models are run with fixed random seeds and deterministic inference settings. 

\section{Extended Results and Analysis}

\subsection{Scenario-Specific Performance Analysis and Qualitative Evaluation}


\begin{table*}[ht]
\centering
\resizebox{0.75\linewidth}{!}{
\begin{tabular}{lccccccc}
\toprule
\textbf{Methods} & $<\delta^{x}_{2}$ $\uparrow$ & $<\delta^{x}_{4}$ $\uparrow$ & $<\delta^{x}_{8}$ $\uparrow$ & $<\delta^{x}_{16}$ $\uparrow$& $<\delta^{x}_{32}$ $\uparrow$& $<\delta^{x}_{avg}$ $\uparrow$& \textbf{EPE} $\downarrow$\\
\midrule
\multicolumn{8}{c}{\textbf{Tissue Deformation}} \\
\midrule
MFT~\cite{neoral2024mft} & 23.35 & \textbf{45.99} & \textbf{73.12} & 81.36 & 92.06 & \underline{63.17} & 24.89 \\
Track-On~\cite{Aydemir2025trackon} & \underline{26.30} & 36.61 & 67.57 & \underline{86.91} & \underline{94.26} & 62.33 & \underline{17.95} \\
TG-SurgPT & \textbf{26.58} & \underline{39.63} & \underline{68.92} & \textbf{87.97} & \textbf{95.21} & \textbf{63.86} & \textbf{15.12} \\
\midrule
\multicolumn{8}{c}{\textbf{Instrument Occlusion}} \\
\midrule
MFT~\cite{neoral2024mft} & 32.53 & \textbf{55.31} & 77.19 & \textbf{91.81} & \underline{95.94} & \underline{70.56} & 17.31 \\
Track-On~\cite{Aydemir2025trackon} & \underline{36.09} & \underline{54.04} & \underline{78.40} & 87.13 & 92.89 & 69.71 & \underline{14.55} \\
TG-SurgPT & \textbf{37.70} & 52.19 & \textbf{79.82} & \underline{89.89} & \textbf{96.38} & \textbf{71.20} & \textbf{10.85} \\
\midrule
\multicolumn{8}{c}{\textbf{Camera Jitter}} \\
\midrule
MFT~\cite{neoral2024mft} & 24.38 & \underline{48.07} & \underline{69.24} & 82.98 & \textbf{92.50} & 63.39 & 25.16 \\
Track-On~\cite{Aydemir2025trackon} & \underline{25.61} & 47.71 & 68.27 & \underline{85.44} & 90.16 & \underline{63.44} & \underline{16.91} \\
TG-SurgPT & \textbf{26.61} & \textbf{49.71} & \textbf{69.77} & \textbf{86.72} & \underline{90.80} & \textbf{64.92} & \textbf{12.85} \\
\midrule
\multicolumn{8}{c}{\textbf{Surface Reflection}} \\
\midrule
MFT~\cite{neoral2024mft} & \underline{24.99} & \underline{48.81} & \textbf{73.83} & \underline{80.66} & \underline{92.90} & \underline{64.28} & 16.36 \\
Track-On~\cite{Aydemir2025trackon} & 24.34 & 48.07 & {67.95} & 80.30 & 91.99 & 63.58 & \underline{14.49} \\
TG-SurgPT & \textbf{25.93} & \textbf{49.76} & \underline{70.06} & \textbf{81.54} & \textbf{94.85} & \textbf{64.90} & \textbf{11.38} \\
\midrule
\multicolumn{8}{c}{\textbf{Cauterization Smoke}} \\
\midrule
MFT~\cite{neoral2024mft} & \underline{30.74} & \textbf{63.20} & 83.15 & 95.88 & \textbf{98.88} & \underline{74.37} & {6.64} \\
Track-On~\cite{Aydemir2025trackon} & 27.11 & 60.45 & \underline{83.55} & \textbf{96.75} & 98.75 & {73.32} & \underline{5.03} \\
TG-SurgPT & \textbf{31.74} & \underline{62.33} & \textbf{84.74} & \underline{96.05} & \underline{98.79} & \textbf{74.73} & \textbf{4.92} \\
\midrule
\multicolumn{8}{c}{\textbf{Mean Results}} \\
\midrule
MFT~\cite{neoral2024mft} & 27.20 & \textbf{52.28} & \underline{75.30} & 86.54 & \underline{94.45} & \underline{67.15} & 18.07 \\
Track-On~\cite{Aydemir2025trackon} & \underline{27.89} & 49.37 & 73.15 & \underline{87.31} & 93.61 & 66.27 & \underline{13.79} \\
TG-SurgPT & \textbf{29.71} & \underline{50.73} & \textbf{75.53} & \textbf{88.03} & \textbf{95.20} & \textbf{67.77} & \textbf{11.02} \\
\bottomrule
\end{tabular}
}
\caption{Scenario-specific tracking performance comparison across five challenging surgical conditions. Results show position accuracy at multiple pixel thresholds ($<\delta^{x}_{k}$ for $k=2,4,8,16,32$), average accuracy ($<\delta^{x}_{avg}$), and endpoint error (EPE in pixels). Bold values indicate best performance, underlined values show second-best results. TG-SurgPT consistently achieves superior performance across all scenarios, demonstrating the effectiveness of text-guided semantic reasoning for surgical point tracking under diverse visual challenges.}
\label{tab:scenario_tracking}
\end{table*}

\begin{figure*}[!h]
  \centering
  \includegraphics[width=0.8\linewidth]{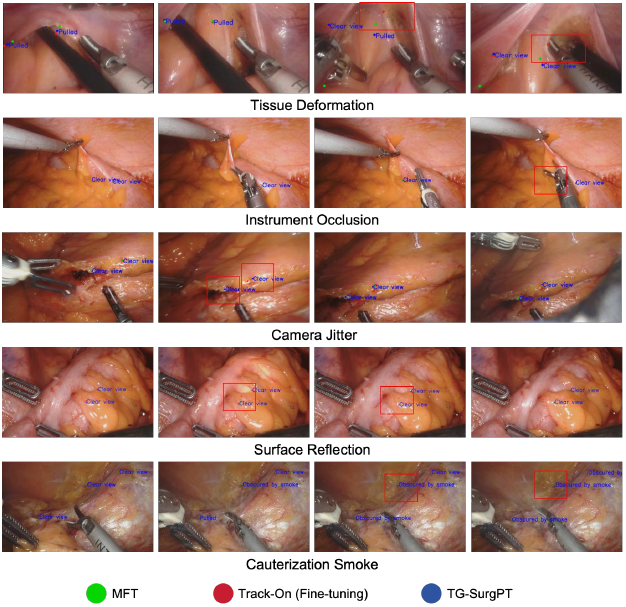}
  \caption{Qualitative comparison of tracking performance across five challenging surgical scenarios. Each row demonstrates a different intraoperative challenge: (1) Tissue Deformation shows progressive anatomical stretching and retraction, (2) Instrument Occlusion illustrates partial obstruction by surgical tools, (3) Camera Jitter displays motion blur and viewpoint instability, (4) Surface Reflection demonstrates specular highlights on moist tissue surfaces, and (5) Cauterization Smoke shows visual degradation from electrosurgical procedures. Point trajectories are color-coded: MFT (green), Track-On (red), and TG-SurgPT (blue). Red bounding boxes highlight regions of particular tracking variance. TG-SurgPT consistently maintains superior spatial accuracy and temporal stability across all scenarios, demonstrating robust performance under diverse surgical conditions.}
\label{fig:tracking_view}
\end{figure*}

\begin{figure*}[!h]
  \centering
  \includegraphics[width=\linewidth]{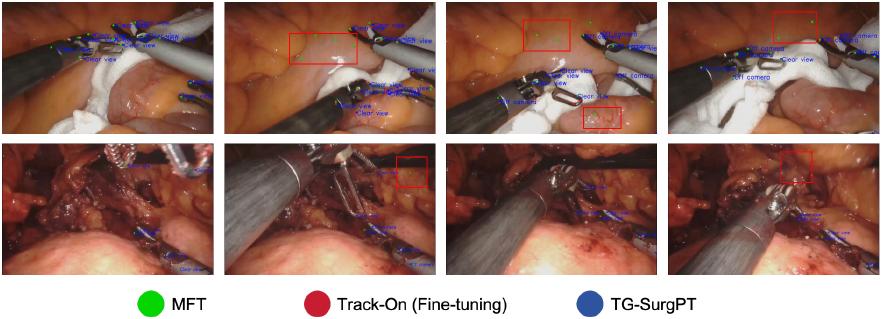}
  \caption{Qualitative comparison of surgical instrument keypoint tracking across challenging visual conditions. Top row demonstrates tracking performance under partial instrument occlusion and tool overlap scenarios. Bottom row illustrates tracking robustness during rapid instrument motion and environmental visual degradation (smoke, blood, debris). Keypoint trajectories are color-coded: MFT (green), Track-On (red), and TG-SurgPT (blue). Red bounding boxes highlight regions of particular tracking difficulty where visual challenges are most pronounced. TG-SurgPT consistently maintains superior keypoint localization accuracy and temporal stability, demonstrating enhanced robustness for surgical point tracking under diverse intraoperative conditions.}
\label{fig:instrument_vis}
\end{figure*}

In this section, we provide a detailed analysis of tracking performance across the five challenging surgical scenarios we categorized, presenting both quantitative results in Table~\ref{tab:scenario_tracking} and qualitative visualizations in Figures~\ref{fig:tracking_view} and~\ref{fig:instrument_vis}. We compare three representative methods: MFT~\cite{neoral2024mft} (optical flow-based vision-only baseline), Track-On~\cite{Aydemir2025trackon} (transformer-based baseline), and our proposed TG-SurgPT (text-guided approach).

\textbf{Tissue Deformation} In the tissue deformation scenario, TG-SurgPT achieves the highest average accuracy ($<\delta^x_{\text{avg}} = 63.86$) and the lowest EPE (15.12), slightly outperforming MFT (green; 63.17, 24.89) and Track-On (red; 62.33, 17.95), as reported in Table~\ref{tab:scenario_tracking}. As visualized in the Tissue Deformation row of Fig.~\ref{fig:tracking_view}, TG-SurgPT (blue markers) remains closely aligned with MFT (green) even as tissue undergoes progressive stretching and reshaping. In contrast, Track-On (red) exhibits increasing drift under mid-to-high deformation levels, particularly in regions with pronounced contour variation. While MFT~\cite{neoral2024mft} predictions remain relatively stable, TG-SurgPT provides enhanced temporal consistency and improved adaptation to non-rigid anatomical motion.

\textbf{Instrument Occlusion} Under occlusion caused by surgical instruments, TG-SurgPT attains the best performance with an average accuracy of 71.20 and an EPE of 10.85, surpassing MFT (green; 70.56, 17.31) and Track-On (red; 69.71, 14.55). As shown in the Instrument Occlusion row of Fig.~\ref{fig:tracking_view}, TG-SurgPT (blue) consistently maintains accurate anatomical localization even when targets are partially or fully obscured by tools such as forceps. Compared to MFT (green), which shows minor instability, Track-On (red) frequently misaligns under occlusion, especially in frames with complex tool-tissue interaction. These results highlight TG-SurgPT’s robustness in resolving transient occlusions and its ability to leverage spatiotemporal continuity for maintaining identity fidelity.

\textbf{Camera Jitter} In scenarios simulating camera jitter and abrupt viewpoint changes, TG-SurgPT again leads in performance, achieving an average accuracy of 64.92 and an EPE of 12.85, outperforming MFT (green; 63.39, 25.16) and Track-On (red; 63.44, 16.91), as shown in Table~\ref{tab:scenario_tracking}. The Camera Jitter row in Fig.~\ref{fig:tracking_view} illustrates TG-SurgPT’s stability under simulated turbulence, such as frame-to-frame motion blur, panning, and tilting. Despite these disturbances, TG-SurgPT (blue) retains consistent alignment with MFT (green), while Track-On (red) displays noticeable drift and jitter-sensitive instability. This demonstrates TG-SurgPT’s capacity to produce temporally stable and spatially precise predictions even under dynamic imaging conditions.

\textbf{Surface Reflection} In the presence of specular reflections, TG-SurgPT achieves an average accuracy of 64.90 and an EPE of 11.38, slightly outperforming MFT (green; 64.28, 16.36) and Track-On (red; 63.58, 14.49). As shown in the Surface Reflection row of Fig.~\ref{fig:tracking_view}, reflections from moist or glossy tissue surfaces challenge localization reliability. TG-SurgPT (blue) maintains precise tracking across reflective regions, closely matching MFT (green), while Track-On (red) exhibits frequent mislocalization, including “jumping” artifacts around highlight boundaries. These results validate TG-SurgPT’s robustness against non-Lambertian distortions common in endoscopic surgery.

\textbf{Cauterization Smoke} In the cauterization smoke scenario, TG-SurgPT delivers the best performance, with an average accuracy of 74.73 and an EPE of 4.92. Although MFT (green) achieves comparable accuracy (74.37), its higher EPE (6.64) indicates reduced localization precision under low-visibility conditions. As depicted in the Cauterization Smoke row of Fig.~\ref{fig:tracking_view}, TG-SurgPT (blue) robustly tracks targets even as dense smoke partially obscures tissue boundaries. In contrast, Track-On (red) frequently fails to maintain stable predictions, exhibiting marker drift or blurring in haze-affected frames. These findings confirm TG-SurgPT’s ability to withstand visual degradation typical of cauterization events in surgical workflows.

\textbf{Overall}, TG-SurgPT achieves the best mean tracking accuracy (67.77) and the lowest mean EPE (11.02) across all five scenarios. Compared with MFT (67.15) and Track-On (66.27), TG-SurgPT demonstrates consistent improvements in both robustness and localization precision. These results validate its adaptability to diverse intraoperative challenges and its suitability for real-world surgical applications.


\subsection{Scenario-Specific Qualitative Tracking Analysis}

\subsubsection{Tissue Subset}
Figure~\ref{fig:tracking_view} presents qualitative tracking results for the tissue subset across five challenging surgical scenarios, comparing our TG-SurgPT method (blue markers) against MFT~\cite{neoral2024mft} (green markers) and Track-On~\cite{Aydemir2025trackon} (red markers) baselines.

\textbf{Tissue Deformation}
As shown in the first row of Fig.~\ref{fig:tracking_view}, \textit{TG-SurgPT} (blue) exhibits stable tracking under progressive anatomical deformation. The blue markers remain tightly aligned with MFT predictions (green) across frames involving tissue stretching and retraction, while Track-On (red) shows noticeable drift, particularly under moderate to severe deformation. This demonstrates TG-SurgPT’s robustness in modeling non-rigid motion, with only occasional deviations during abrupt release events.

\textbf{Surface Reflection}
In the presence of specular highlights (second row of Fig.~\ref{fig:tracking_view}), TG-SurgPT effectively mitigates photometric distortions caused by moist or glossy tissue surfaces. Blue markers maintain consistent localization across reflective boundaries, whereas Track-On (red) frequently jumps to visually similar but incorrect regions. While MFT (green) remains reasonably stable, TG-SurgPT further enhances robustness in challenging lighting conditions by preserving point-level accuracy under non-Lambertian reflections.

\textbf{Instrument Occlusion}
The third row of Fig.~\ref{fig:tracking_view} illustrates performance under partial occlusion caused by tool-tissue and tool-tool interactions. TG-SurgPT (blue) successfully preserves identity and localization of occluded instrument joints and tips, maintaining coherent trajectories even as tools overlap. In contrast, Track-On (red) frequently misassociates keypoints, leading to identity swaps or merging of adjacent structures. MFT (green) offers relatively consistent tracking, though TG-SurgPT exhibits superior disambiguation in occlusion-heavy frames.

\textbf{Camera Jitter}
Under simulated camera jitter and abrupt endoscope motion (fourth row of Fig.~\ref{fig:tracking_view}), TG-SurgPT delivers temporally stable predictions. Blue markers remain aligned with MFT outputs (green) despite rapid viewpoint shifts or blur, while Track-On (red) suffers from erratic motion and delayed re-alignment. This highlights TG-SurgPT’s ability to suppress jitter-induced noise and maintain continuity across motion bursts.

\textbf{Cauterization Smoke}
In the fifth row of Fig.~\ref{fig:tracking_view}, TG-SurgPT demonstrates strong resilience to visual degradation caused by cauterization smoke. Despite dense haze and blurred boundaries, blue markers remain stably anchored to anatomical structures, closely following MFT predictions (green). In contrast, Track-On (red) often loses alignment, with markers drifting or fluctuating in dense smoke regions. These results indicate TG-SurgPT’s capability to sustain accurate predictions even under severe low-visibility conditions typical of electrosurgical procedures.

\subsubsection{Instrument Subset}

Figure~\ref{fig:instrument_vis} illustrates representative examples of marker tracking under occlusion, motion, and environmental noise. Across all scenarios, TG-SurgPT (blue) consistently demonstrates higher spatial and temporal fidelity compared to Track-On (red) and MFT (green). Under partial occlusion such as forceps crossing, TG-SurgPT accurately retains marker positions on instrument joints, while Track-On frequently misassigns them to nearby tissue or overlapping tools. MFT~\cite{neoral2024mft} shows moderate robustness but exhibits more deviations than TG-SurgP. During fast instrument motion, TG-SurgPT maintains precise alignment with MFT and offers improved stability, whereas Track-On~\cite{Aydemir2025trackon} displays noticeable drift or lag. In frames with visual clutter, including smoke, debris, or blood, TG-SurgPT continues to prioritize instrument structure and avoids incorrect associations. In contrast, Track-On~\cite{Aydemir2025trackon} often misplaces markers, and MFT~\cite{neoral2024mft} occasionally shows instability.

These observations confirm that TG-SurgPT effectively handles real-world surgical challenges such as occlusion, high-speed movement, and intraoperative noise. The model preserves instrument identity across diverse conditions, supporting its applicability to critical tasks like robotic tool tracking and intraoperative decision support.

\end{document}